\documentclass[letterpaper, 10 pt, conference]{ieeeconf}  

\IEEEoverridecommandlockouts                              

\usepackage{graphicx} 
\usepackage{amsmath} 
\usepackage{amssymb}  
\usepackage{xcolor}
\usepackage{booktabs}
\usepackage{multirow}
\usepackage{subcaption}
\usepackage{balance}
\usepackage{cite}

\DeclareMathOperator*{\argmax}{arg\,max}

\title{\LARGE \bf
Drone Soccer: Learning to Manipulate with Multicopter Downwash}

\author{Neelay Joglekar$^{1}$, Bavin Saravanan$^{1}$, Yutong Wang$^{1}$, Varun Kandiyappan$^{1}$,\\ Junyi Geng$^{2}$, and Sebastian Scherer$^{1}$%
\thanks{This work was supported in part by the Office of Naval Research under Grant N000142512232 and in part by the National Science Foundation Graduate Research Fellowship Program under Grant 2140739 and 2631988. Any opinions, findings, and conclusions or recommendations expressed in this material are those of the author(s) and do not necessarily reflect the views of the National Science Foundation.}%
\thanks{$^{1}$ Neelay Joglekar, Bavin Saravanan, Yutong Wang, Varun Kandiyappan, and Sebastian Scherer are with The Robotics Institute, School of Computer Science, Carnegie Mellon University, Pittsburgh, PA 15213, USA
    {\tt\small \{njogleka, bsaravan, yutongw3, dkandiya, basti\}@andrew.cmu.edu}.}%
\thanks{$^2$ Junyi Geng is with the Department of Aerospace Engineering, Pennsylvania State University, University Park, PA 16802, USA
    {\tt\small jgeng@psu.edu}.}%
}

\begin{document}

\maketitle
\thispagestyle{empty}
\pagestyle{empty}

\begin{abstract}

Although multicopter drones are traditionally designed for “perception-only” tasks, like mapping and exploration, recent work has sought to develop Unmanned Aerial Manipulators (UAMs) to solve mobile manipulation tasks. Aerial manipulation performance can be impacted by ``downwash," the airflow produced by propellers, but current state-of-the-art UAMs either ignore downwash or treat it as a disturbance. Instead, is it possible to actively use downwash as a tool during manipulation? We design a drone soccer task to explore the feasibility of downwash-based manipulation. Specifically, we develop a simplified downwash dynamics model which we use to train an RL policy to dribble a soccer ball. We further demonstrate that our policy transfers to real world deployment. This work provides key insights into novel manipulation capabilities for multicopters.

\end{abstract}

\section{INTRODUCTION}

Although multicopter drones are traditionally designed for “perception-only” tasks, like mapping and exploration, recent work has sought to develop Unmanned Aerial Manipulators (UAMs) to solve mobile manipulation tasks. Often, UAM drones utilize direct contact to perform manipulation, sporting arms \cite{lee2021aerial, liang2022adaptive, guo2024flying, he2025flying, gupta2025umi, zhan2026contact-aware}, cables \cite{geng2020cooperative, geng2022load, sun2025agile}, and other tools \cite{hehn2011a, muller2011quadrocopter} to change lightbulbs in hard-to-reach areas \cite{he2025flying, gupta2025umi}, carry loads \cite{geng2020cooperative, geng2022load, sun2025agile}, and play racket sports \cite{muller2011quadrocopter}. However, is physical contact the only interaction mode available for UAMs?

In particular, drones constantly interact with their surroundings via ``downwash," the airflow field produced by drone's propellers. Prior aerial manipulation works either ignore downwash or treat it as a disturbance \cite{cao2025proximal}. However, downwash could also be a vital tool for solving novel manipulation tasks. For example, pesticide drones passively rely on downwash to penetrate the leaf canopy, allowing the pesticide treatement to cover a larger plant surface area \cite{zhu2022cfd}. Taking this one step further, is it possible to actively use downwash as a tool during manipulation?

\begin{figure}[t]
    \centering
    \includegraphics[width=\linewidth]{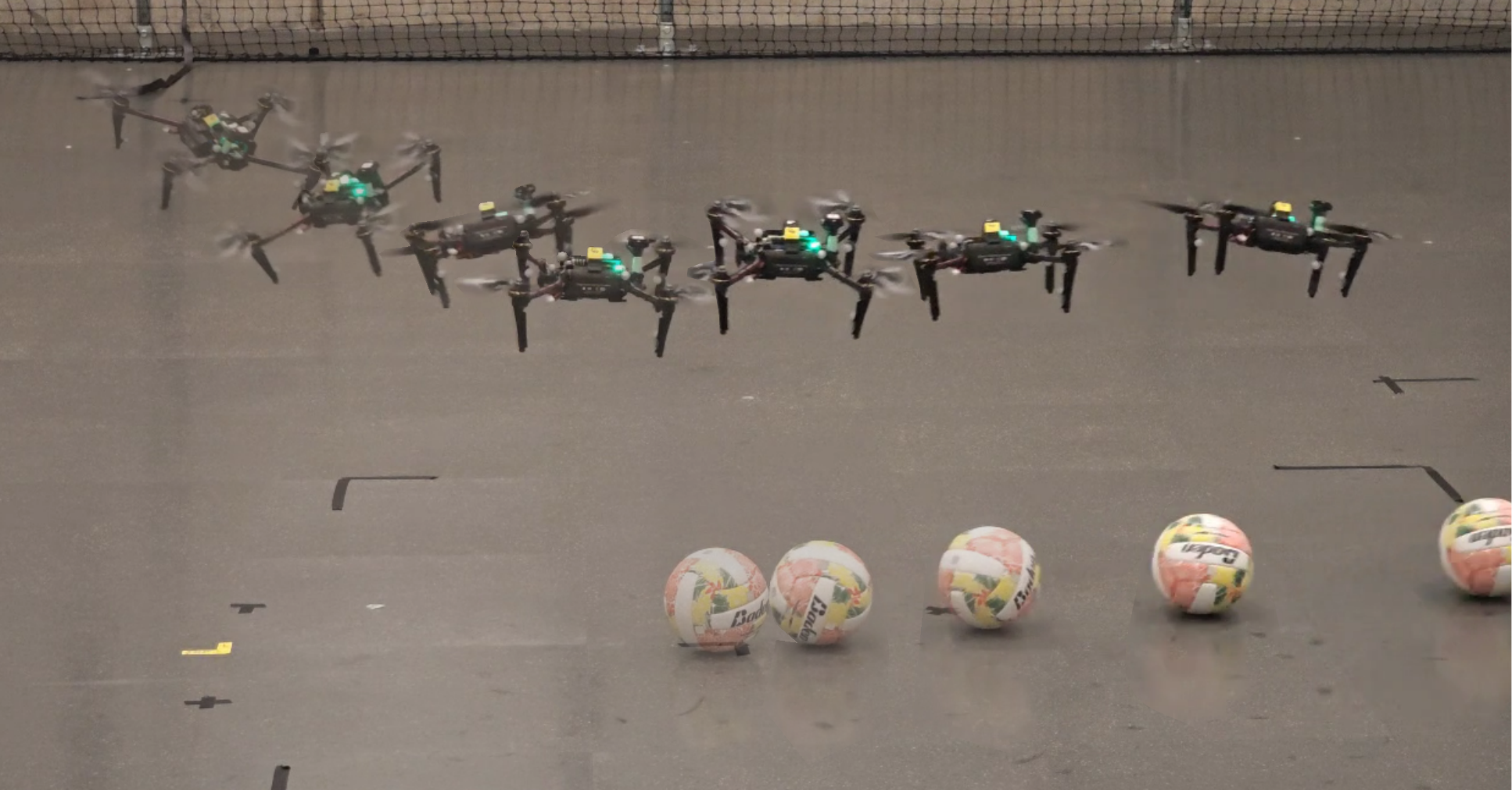}
    \caption{Aerial manipulation seeks to equip drones to with the capability to solve mobile manipulation tasks. Prior work utilizes direct physical contact for manipulation, but we show that drones can use the ``downwash" airflow produced from their propellers to manipulate objects. We specifically develop a closed-loop RL policy, trained using a simplified downwash model, to solve a ``drone soccer" task.}
    \label{title}
\end{figure}

Prior work in ``pneumatic manipulation" suggests this is possible \cite{becker2009automated, xu2022dextairity}. Pneumatic manipulation has been applied in several industrial settings \cite{laurent2015survey}, including using a vacuum attachment on a robotic arm for depowdering during additive manufacturing \cite{liu2022robotic}. Wu et al. introduced pneumatic mobile manipulation, attaching a blower to a wheeled mobile robot and training a policy to blow scattered particles into a collection receptacle \cite{wu2022learning}. For this task, pneumatic manipulation was more effective than non-prehensile pushing. Nevertheless, no prior work has explored pneumatic manipulation with multicopter downwash.

``Downwash-based" manipulation is uniquely challenging compared to other pneumatic manipulation tasks. Modeling downwash is incredibly difficult, as it varies widely with the drone’s highly-dynamic motion, proximity to surfaces/objects, etc. In most cases, downwash can only be accurately modeled with computationally intensive continuous fluid dynamics (CFD) methods \cite{zhu2022cfd}, which are intractable for use in both MPC and RL-based control. There exist simpler downwash models, but these are only valid for stationary drones with no nearby surfaces \cite{bauersfeld2025robotics}.

Nevertheless, closed-loop policies have shown clear promise in enabling effective control and manipulation under complex dynamics, even when paired with inaccurate models. Drone dynamics exemplify this, as simple linear models that ignore aerodynamic and some inertial effects are often enough to enable agile flight via MPC \cite{nguyen2024tinympc}. Furthermore, Wu et al. showed that a ``multi-frequency" hierarchical closed loop design directly transferred from sim-to-real, even when trained with simplified particle-based fluid dynamics \cite{wu2022learning}. In a similar manner, can simple downwash models paired with closed-loop control enable downwash-based manipulation?

\subsection{Our Contribution}

In this manuscript, we design and demonstrate a drone soccer task to explore the feasibility of downwash-based manipulation. Our specific contributions are as follows:
\begin{itemize}
    \item We develop the first-ever UAM control policy that actively utilizes downwash to perform manipulation;
    \item We design a simple downwash dynamics model that mimics fluid-ground interaction effects;
    \item We demonstrate sim-to-real transfer of a closed-loop RL policy trained with our downwash model.
\end{itemize}
Our work provides key insights into novel manipulation capabilities for multicopters.

\section{METHODS}

We define the drone soccer task as follows: given a goal position, the drone must push a ball at high velocity towards a goal position; the drone can only manipulate the ball using downwash, not direct contact. For a policy to solve this task, it must reason about downwash dynamics. Hence, we first develop a downwash dynamics model to translate drone actions into forces on the ball. Then, we use this model to train an RL drone soccer policy in simulation.

\begin{figure}[t]
    \centering
    \includegraphics[width=\linewidth]{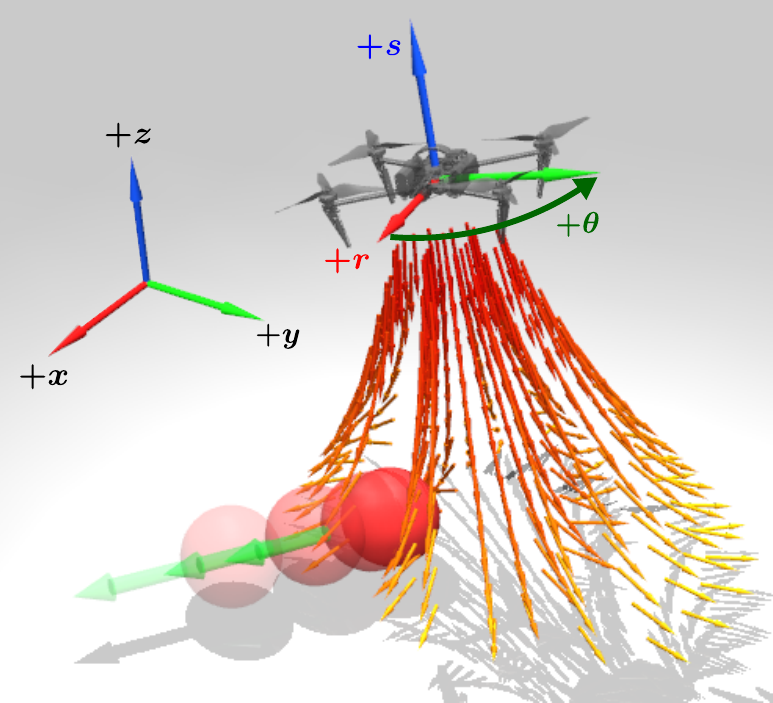}
    \caption{Our downwash model is defined in the mixed-polar local frame of the drone. Our downwash field re-orients with drone roll-pitch and pushes radially outwards as it approaches the ground. We use a blunt drag model to convert fluid velocity to force on the ball. When trained with this simulated model, our RL policy transfers well to real deployment.}
    \label{downwash}
\end{figure}

\subsection{Downwash Model}

As visualized in Fig. \ref{downwash}, our downwash model takes a motor thrust action $u \in \mathbb{R}^N$, where $N$ is the number of propellers, and converts it into an airflow velocity field. The velocity field is defined w.r.t. the drone's local frame in mixed polar coordinates, where the $s$-axis is orthogonal to the rotor plane, $r$ is the radial axis within the rotor plane, and $\theta$ defines the orientation of $r$ about $s$. The world frame is level with the ground, with $z$-axis pointing up.

Our model first converts $u$ to an estimated ``induced velocity" $U_H$, which is the flow-aligned airspeed produced at the rotor plane. We average the $N$ propeller thrusts and use Momentum Theory to estimate $U_H$ as follows \cite{leishman2006principles}:
\begin{equation} \label{Uh}
    U_H = \sqrt{\frac{\boldsymbol{1} \cdot u}{2\rho A_p N}}
\end{equation}
where $\rho$ is air density and $A_p$ is the disk area of a single propeller. Eq. \eqref{Uh} ignores induced velocity variation between the $N$ propellers and different drone motion profiles, but we find it is sufficient for drone soccer.

Given $U_H$, we utilize Baursfeld et al.'s simplified downwash model \cite{bauersfeld2025robotics} to estimate a downwash speed:
\begin{align}
    \tilde{s} &= \frac{s}{l}, \quad \tilde{r} = \frac{r}{l}, \quad \xi = \frac{\tilde{r}}{S(\tilde{s} - s_0)}\label{normalization}\\
    U &= \frac{U_HBd}{(\tilde{s} - s_0)(1 + (\sqrt{2}-1)\xi^2)^2} \label{U}
\end{align}
where $\xi$ is the rescaled radial position associated with $r$ and $Bd = 10.11, ~S=0.07668, ~s_0 = -5.817$ are fitted parameters computed from extensive anemometer data. Eqns. \eqref{normalization}-\eqref{U} ignore radial flow components \cite{pope2000turbulent}, but these velocities are small enough to be assumed negligible.

In addition to $U$, we need to model the downwash velocity direction $\vec v(s, r, \theta)$. The downwash velocity field radiates outward as it approaches the ground \cite{barata2004laser}. We can heuristically approximate this behavior by defining a ``spreading angle" $\phi$:
\begin{align}
    \phi &= \arctan \frac{r}{e_3\cdot \mathbf{T} \begin{bmatrix}
        rcos\theta & r\sin\theta & s
    \end{bmatrix}^\top} \label{phi}\\
    \vec{v} &= \begin{bmatrix}
        \sin{\phi}\cos{\theta}& \sin{\phi}\sin{\theta} & -\cos{\phi}
    \end{bmatrix}^\top \label{vecv}
\end{align}
where $e_3 = \begin{bmatrix}
    0 & 0 & 1
\end{bmatrix}^\top$, $\mathbf{T} \in SE(3)$ is local-to-world transformation matrix, and $\vec{v}$ is a unit vector. Intuitively, $\vec{v}$ aligns with the $s$-axis near the rotor plane origin and gradually orients outwards near the ground. With $U$ and $\vec{v}$ defined, we can compute downwash velocity $v(r, s, \theta)$ as:
\begin{equation}
    v = U\mathbf{R}\vec{v}
\end{equation}
where $\mathbf{R} \in SO(3)$ is the local-to-world rotation matrix.

Lastly, we estimate downwash force on the ball. Upon computing $v$ at the ball's position, we use the following blunt drag equation \cite{munson1990fundamentals}.
\begin{equation} \label{drag}
    F_b = C_d\rho A_b \|v - v_b\|(v - v_b)
\end{equation}
where $C_d = 0.5$ is the drag coefficient, $A_b$ is the ball's cross-sectional area, $v_b$ is the current ball velocity, and $F_b$ is the drag force. Eqn. \eqref{drag} doesn't properly capture the effects of nonuniform, turbulent downwash on the ball, but we find it is sufficient for drone soccer.

Hence, \eqref{Uh}-\eqref{drag} compose our downwash dynamics model. As mentioned, our model is built on several assumptions and heuristics. We handle resulting dynamics inaccuracies with a closed-loop controller.

\subsection{RL Policy}

We train a reinforcement learning policy from our downwash model to perform drone soccer. We represent the drone soccer task as a Markov Decision Process $(\mathcal{S}, \mathcal{A}, \mathcal{P}, \mathcal{R})$, with state space $\mathcal{S}$, action space $\mathcal{A}$, stochastic transition function $\mathcal{P} : \mathcal{S} \times \mathcal{A} \times\mathcal{S} \to \mathbb{R}^+$, and reward function $\mathcal{R} : \mathcal{S} \times \mathcal{A} \times \mathcal{S} \to \mathbb{R}$. This representation allows us to define a policy $\pi : \mathcal{S} \to \mathcal{A}$ with an associated value function:
\begin{equation} \label{value}
    V^\pi(s) =  \mathbb{E}_{s' \sim \mathcal{P}(s'|s, \pi(s))}[r(s,\pi(s),s') + \gamma V^\pi(s')]
\end{equation}
where $s, s' \in \mathcal{S}$, and $\gamma \in [0, 1]$ is the discount factor. The optimal drone soccer policy satisfies the following condition:
\begin{equation} \label{policy}
    \pi(s) = \argmax_{a\in\mathcal{A}} ~ \mathbb{E}_{s' \sim \mathcal{P}(s'|s, a)}[r(s,a,s') + \gamma V^\pi(s')]
\end{equation}
Towards this end, we first define $\mathcal{S} \subset \mathbb{R}^{18}$ to encapsulate all relevant drone, ball, and goal configurations. Each $s \in \mathcal{S}$ is defined as:
\begin{equation}\label{s}
    s = \begin{bmatrix}
        z_m & \vec{s}^\top & v_m^\top & \omega_m^\top & (p_b - p_m)^\top & v_b^\top & (g - \begin{bmatrix}
            x_b \\ y_b
        \end{bmatrix})^\top
    \end{bmatrix}
\end{equation}
where: $z_m \in \mathbb{R}^1$ is drone world frame height; $\vec{s} \in \mathbb{R}^3$ is the drone body $s$-axis unit vector; $v_m, \omega_m \in \mathbb{R}^3$ are drone world-frame linear and body-frame angular velocities; $p_b, p_m \in \mathbb{R}^3$ are the drone and ball world-frame positions; $v_b \in \mathbb{R}^3$ is the ball linear velocity; $g \in \mathbb{R}^2$ is the 2D goal position; $x_m, y_m$ are the $xy$ coordinates of $p_b$. Our state representation ignores drone yaw because our downwash model doesn't depend on yaw. Note that we represent $xy$ positions of the drone, ball, and goal as relative w.r.t. each other.

We represent each action $a \in \mathcal{A} \subset \mathbb{R}^3$ as a mixed-frame waypoint:
\begin{equation}\label{a}
    a = \begin{bmatrix}
        \Delta x_m & \Delta y_m & z_m
    \end{bmatrix}
\end{equation}
where $\Delta x_m, \Delta y_m$ are displacements from the current drone position in the world frame. This relative representation is consistent with our state space structure. Our policy supplies waypoints at 10 Hz to a low-level controller, which produces the motor thrust vector $u$.

We define $\mathcal{R}$ as:
\begin{equation}\label{reward}
\begin{aligned}
    \mathcal{R}(s, a, s') &= 
    w_v\frac{s'_{17:18}}{\|s'_{17:18}\|} \cdot s'_{14:15} + w_p(G(s'_{17:18}) - G(s_{17:18})) \\
    &+ w_r H(\epsilon_g - \|s'_{17:18}\|) - w_cH(\epsilon_z - s'_1)\\
    &- w_cH(\epsilon_b - \|s'_{12:13}\|) - w_s L(a)
\end{aligned}
\end{equation}
where $s_{i:j}$ indexes \eqref{s}, $G$ is a 2D gaussian centered on $g$, $H$ is the heavyside function, and $w_{(\cdot)}$ are weights. The first two terms encourage the policy to push the ball towards the goal. The third term is a sparse reward that activates when the ball is within $\epsilon_g$ distance of the goal. The fourth and fifth terms are crash penalties that activate when the drone is within $\epsilon_z$ distance to the ground or $\epsilon_b$ distance to the ball, respectively. Lastly, we add a waypoint smoothness cost defined as follows:
\begin{equation}\label{smoothness}
    L(a) = \|(p_m + a) - (p_{m, prev} + a_{prev})\|
\end{equation}
where $p_{m,prev}, a_{prev}$ are the drone state and action from the previous timestep. This penalty discourages large changes in absolute waypoint position.

With our MDP defined, we can set up a simulation environment and find an approximate solution to \eqref{policy}. Our simulation setup is detailed in Section \ref{sec:exp}. We use PPO \cite{schulman2017proximal} to train $\pi$, designing a training curriculum to gradually increase episode difficulty. Early in training, the drone is initialized behind the ball, so the policy quickly learns to push the ball. Later, initial position of the drone, ball, and goal are completely randomized. Additionally, early in training the ball always initializes with zero velocity, but as training progresses it initializes with random velocities of increasing max magnitude.

\section{EXPERIMENTS \& RESULTS} \label{sec:exp}



\begin{figure*}[t]
  \centering
  \begin{minipage}[c]{0.48\textwidth}
    \centering
    \begin{subfigure}[b]{0.48\linewidth}
      \centering
      \includegraphics[width=\linewidth]{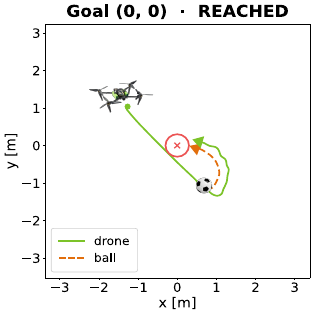}
      \caption{}
      \label{exp:near}
    \end{subfigure}\hfill
    \begin{subfigure}[b]{0.48\linewidth}
      \centering
      \includegraphics[width=\linewidth]{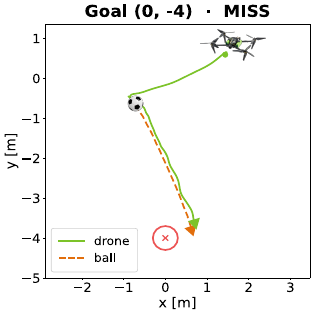}
      \caption{}
      \label{exp:far}
    \end{subfigure}

    \vspace{0.6em}

    \begin{subfigure}[b]{0.48\linewidth}
      \centering
      \includegraphics[width=\linewidth]{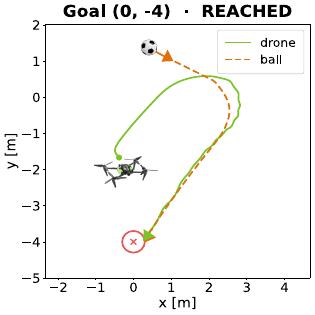}
      \caption{}
      \label{exp:guidance}
    \end{subfigure}\hfill
    \begin{subfigure}[b]{0.48\linewidth}
      \centering
      \includegraphics[width=\linewidth]{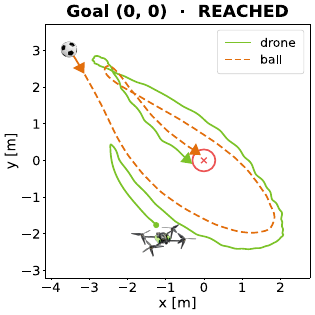}
      \caption{}
      \label{exp:redirect}
    \end{subfigure}
  \end{minipage}%
  \hfill
  \begin{subfigure}[c]{0.48\textwidth}
    \centering
    \includegraphics[width=\linewidth]{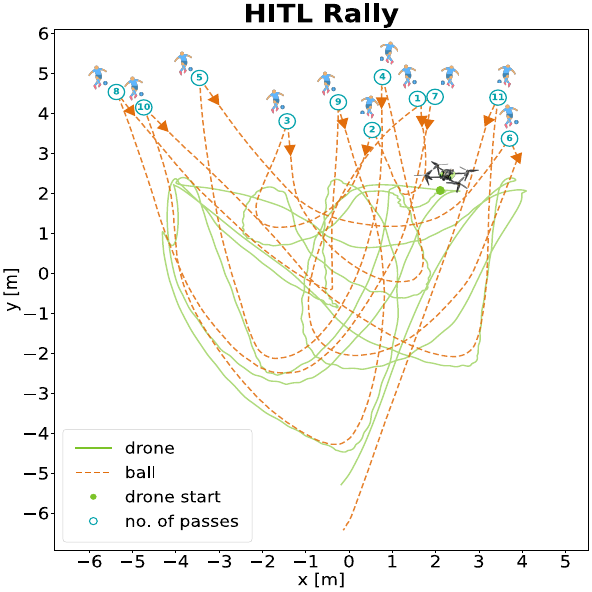}
    \caption{}
    \label{hitl}
  \end{subfigure}

  \caption{(a--d) Geofenced experiments across various drone, ball, and goal configurations. The policy successfully directed the ball to the goal in 3 of 4 configurations; insufficient force occasionally leads to slower convergence (d) or failure (b). (e) Human-in-the-loop (HITL) rally experiment where the drone passes to a goal outside the geofence and the human kicks it back. The pair completed 10 consecutive passes before failing on the 11th due to excessive ball speed.}
  \label{fig:all-experiments}
\end{figure*}

\subsection{Implementation \& Environment Setup}

We use a 2-layer MLP with hidden dimension 64 to represent $\pi$. We bound our action space within $\Delta x_m, \Delta y_m \in [-3, 3]$ meters and $z_m \in [0.5, 1.5]$ meters. Our reward weights are: $w_v = 2$, $w_p = 0.5$ (with gaussian standard deviation 0.1m), $w_r = 500$, $w_c = 100$, and $w_s = 0.005$. Additionally, we set $\epsilon_g = 0.3, \epsilon_b = 0.3, \epsilon_g = 0.2$ meters. Since our simulation environment runs at a higher frequency than our policy, we compute \eqref{reward} at each sim step and input the summed reward between policy inferences to PPO.

We deployed our policy on a Modal AI Starling 2 Max drone. We used a small ball with mass of 148g (about a fourth of the drone's weight) and a diameter of $13.5$ cm. We attached markers to the drone and reflective stickers to the ball so they could be tracked with motion capture. We estimated drone velocity with the PX4 EKF \cite{meier2015px4} and ball velocity with a separate Kalman filter.

We configured a training environment in MuJoCo \cite{todorov2012mujoco} to match our hardware setup. We matched the drone and ball's inertial parameters from measurements and product specs. We mimicked PX4 controller performance using a geometry controller \cite{lee2010geometric}, with proper gains copied from the drone's onboard configuration. Additionally, we found that setting $s_0$ in \eqref{normalization} to $-18.07$ produced ball-downwash dynamics that closely mimicked our collected data. For safety during deployment, we defined a geofence within the $8m \times 10m$ motion capture workspace that prevented the RL policy from commanding waypoints outside a safe region.

\subsection{Hardware Experiments}

We set up 5 hardware experiments to probe policy performance under varying conditions, as visualized in Fig. \ref{fig:all-experiments}. For the first 4 experiments, we initialize the drone, ball, and goal within the geofence and roll out our policy. The ball is initialized with zero velocity for the first 2 experiments (Fig. \ref{exp:near}, \ref{exp:far}) and with nonzero velocity for the next 2 experiments (Fig. \ref{exp:guidance}, \ref{exp:redirect}). In the 5th experiment (Fig. \ref{hitl}), we shrink the goefence to end at $y=+2$ and initialize the goal outside at $(0, 4)$. Hence, the policy is incentivized to push the ball outside the geofence, where a human is waiting to kick the ball back into the drone's workspace. This results in a human-in-the-loop (HITL) experiment where the human and drone continue to pass the ball between each other for as long as possible.

Qualitatively, the policy can effectively manipulate the soccer ball in most scenarios. In particular, the drone achieves 3/4 of the goals within the geofence and completes 10 passes in the HITL experiments. However, the policy sometimes cannot provide enough force to quickly redirect the ball. This is clear in Fig. \ref{exp:far} where the drone encroaches upon the geofence before it can redirect the ball to the goal, and in Fig. \ref{exp:redirect} where the ball initially misses the goal. This is likely a consequence of the sim-to-real gap. According to Tables \ref{table_geo}, \ref{table_hitl}, the policy performs well when the ball's speed stays below $\approx 1.5$m/s, which explains why the policy couldn't handle the final $2.52$ m/s kick in the HITL experiment.

\begin{table}[t]
\caption{Geofenced Experiment Performance}
\label{table_geo}
\begin{center}
\begin{tabular}{c|c|c|c|c}
\toprule
 & (a) & (b) & (c) & (d)\\
\hline
Goal Reached? & True & False & True & True\\
Median Ball Speed (m/s) & 0.43 & 0.50 & 0.90 & 0.91\\
Initial Ball Speed (m/s) & 0.0 & 0.0 & 0.98 & 1.63\\
Final Ball Speed (m/s) & 0.72 & 1.41 & 1.60 & 1.20\\
\bottomrule
\end{tabular}
\end{center}
\end{table}

\begin{table}[t]
\caption{HITL Average Ball Speed Statistics (m/s)}
\label{table_hitl}
\begin{center}
\begin{tabular}{c|c|c|c}
\toprule
Min & Median & Max Successful & Max\\
\hline
0.62 & 1.22 & 1.52 & 2.52\\
\bottomrule
\end{tabular}
\end{center}
\end{table}

\section{DISCUSSION \& CONCLUSION}

We demonstrate that downwash can be used for manipulation. Our simple, heuristic downwash model is sufficient to train an RL policy that transfers well to real world drone soccer. Nevertheless, manipulation performance could still be improved, possibly through real world reinforcement learning \cite{pan2026learning}.

The drone soccer task can be extended to provide insights for additional tasks. Drones could be applied to farm animal herding, which despite having different dynamics still has the same problem formulation as drone soccer. Similar to Wu et al., our work could be extended to solving multi-particle collection tasks, like leaf blowing and litter collection. Another natural extension is multi-agent drone soccer, which can provide insights into collaborative swarm strategy with higher degrees of freedom than traditional robot soccer works. An adversarial drone soccer game can provide insights into swarm-on-swarm strategy, an important concept for defense.





\bibliographystyle{IEEEtran}
\balance
\bibliography{root}

\end{document}